\documentclass[lettersize,journal]{IEEEtran}
\usepackage{amsmath,amsfonts}
\usepackage{algorithmic}
\usepackage{array}
\usepackage{textcomp}
\usepackage{stfloats}
\usepackage{url}
\usepackage{verbatim}
\usepackage{graphicx}
\usepackage[dvipsnames]{xcolor}
\usepackage{subfigure}
\usepackage{booktabs}
\usepackage{breqn}
\usepackage{ulem}
\usepackage{gensymb}
\usepackage{float}

\def\BibTeX{{\rm B\kern-.05em{\sc i\kern-.025em b}\kern-.08em
    T\kern-.1667em\lower.7ex\hbox{E}\kern-.125emX}}
\usepackage{balance}
\begin{document}
	
\title{Reinforcement Learning based Voice Interaction to Clear Path for Robots in Elevator Environment}

\author{Wanli Ma$^\ddagger$$^1$,Xinyi Gao$^\dagger$$^1$,Jianwei Shi$^\star$$^1$,Hao Hu$^\dagger$$^*$,Chaoyang Wang$^\dagger$,Yanxue Liang$^\dagger$, Oktay Karakus$^\lozenge$
	
\thanks{$^\ddagger$Wanli Ma was with the Center of Innovative Research, Westlake University, Hangzhou, PR China. He is now with the School of Computer Science and Informatics, Cardiff University, UK.}

\thanks{$^\star$Jianwei Shi is with the School of Mechanical and Power Engineering, Zhengzhou University, Zhengzhou 450001, China.}

\thanks{$^1$These authors contributed equally to this paper and should be considered co-first authors.}

\thanks{$^\lozenge$Oktay Karakus is with the School of Computer Science and Informatics, Cardiff University, UK.}

\thanks{$^\dagger$Authors are with Center of Innovative Research, Westlake University, Hangzhou, PR China.$^*$Corresponding Author. Email:{huhaoseu@gmail.com.}}

}

\markboth{}
{Reinforcement Learning based Voice Interaction to Clear Path for Robots in Elevator Environment}

\maketitle

\begin{abstract}
Efficient use of the space in an elevator is very necessary for a service robot, due to the need for reducing the amount of time caused by waiting for the next elevator. To provide a solution for this, we propose a hybrid approach that combines reinforcement learning (RL) with voice interaction for robot navigation in the scene of entering the elevator. RL provides robots with a high exploration ability to find a new clear path to enter the elevator compared to traditional navigation methods such as Optimal Reciprocal Collision Avoidance (ORCA). The proposed method allows the robot to take an active clear path action towards the elevator whilst a crowd of people stands at the entrance of the elevator wherein there are still lots of space. This is done by embedding a clear path action (voice prompt) into the RL framework, and the proposed navigation policy helps the robot to finish tasks efficiently and safely. Our model approach provides a great improvement in the success rate and reward of entering the elevator compared to state-of-the-art navigation policies without active clear path operation.
\end{abstract}



\section{Introduction}

\IEEEPARstart{M}{obile} robots have become a research hotspot for various applications due to the critical impact of the Novel Coronavirus Pneumonia~\cite{vongbunyong2020simulation}. Particularly, cleaning and disinfection, security, logistics, and catering delivery are just some of the uses for mobile robots, which have also constructed a line of defence for humans caught up in the epidemic. Although mobile robot navigation capabilities have vastly increased recently, there are still numerous issues in real-world applications, such as poor route flexibility, and time consumption for delivery robots. Specifically, delivery robots working in a multi-floor environment must use an elevator to complete the duty of delivering goods or merchandise. To make this process safe and efficient, there is still lots of room that can be improved. When it comes to safety navigation, there are many navigation algorithms that have been proposed to guarantee that the moving robots are controlled to avoid collection with people or obstacles. However, how to improve the efficiency of the delivery robots on the basis of ensuring safety still requires more efficient solutions. In a multi-layer environment, such as a hotel, one of the vital factors for improving the efficiency of mobile robots is to reduce waiting time for elevators. Nowadays, most navigation methods for robot entering the elevator is offline \cite{sahloul2021hybridization, shah2021ving}, which creates a fixed route first and then controls the robot running along the route. Although offline navigation could work in the current situation and be able to generate a new path for the robot to enter the elevator only in some specific scenes, e.g. its route is blocked by humans or other obstacles. However, due to the environment always changing with human movement, offline navigation is likely to lose the optimal path at some moments. On the contrary, online navigation can change the path to be optimal in real-time according to the current condition, which is obviously more flexible and efficient than offline mode but requires high computing power. In the current circumstances,  with the increase in computing power of computing devices, online navigation algorithms have developed rapidly, but the literature and industry are still far from fully leveraging the advantages of combining online navigation and voice interaction to improve the efficiency of delivery robots in multi-layer environments. Therefore, this paper aims to provide an online navigation method for delivery robots to enter the elevator safely and efficiently.

When confronted with a complicated and changing environment, people can usually adopt the best navigation technique and get through safely and quickly. In particular, in cases when we are in a hurry to get things done, for example, when efficiency is our priority, we would frequently change our path to be the shortest path and also remind others around to make room for us so that we can reach our destination as soon as possible. Although a few mobile robot navigation systems focus on efficient navigation, they use human senses in specific scenes to remind pedestrians to give way and improve the robot's navigation efficiency~\cite{3,4,5,6,7}. However, those methods realize this robot-human interaction by pre-configured and rigid settings, which is difficult to cover all situations that need robot-human interaction. Also, in simulations, ambitious and highly frequent warnings present an unnatural behaviour and might make other agents change their original moving direction and make their behaviour disordered. This will clearly lead to an increased likelihood of collisions. \cite{Nishimura2020L2BLT}.

\begin{figure}[t]
	\centering
\setlength{\abovecaptionskip}{0.cm}
\setlength{\belowcaptionskip}{-0.cm}
\setlength{\dbltextfloatsep}{0pt}
\subfigure[]{\includegraphics[width=3.5cm]{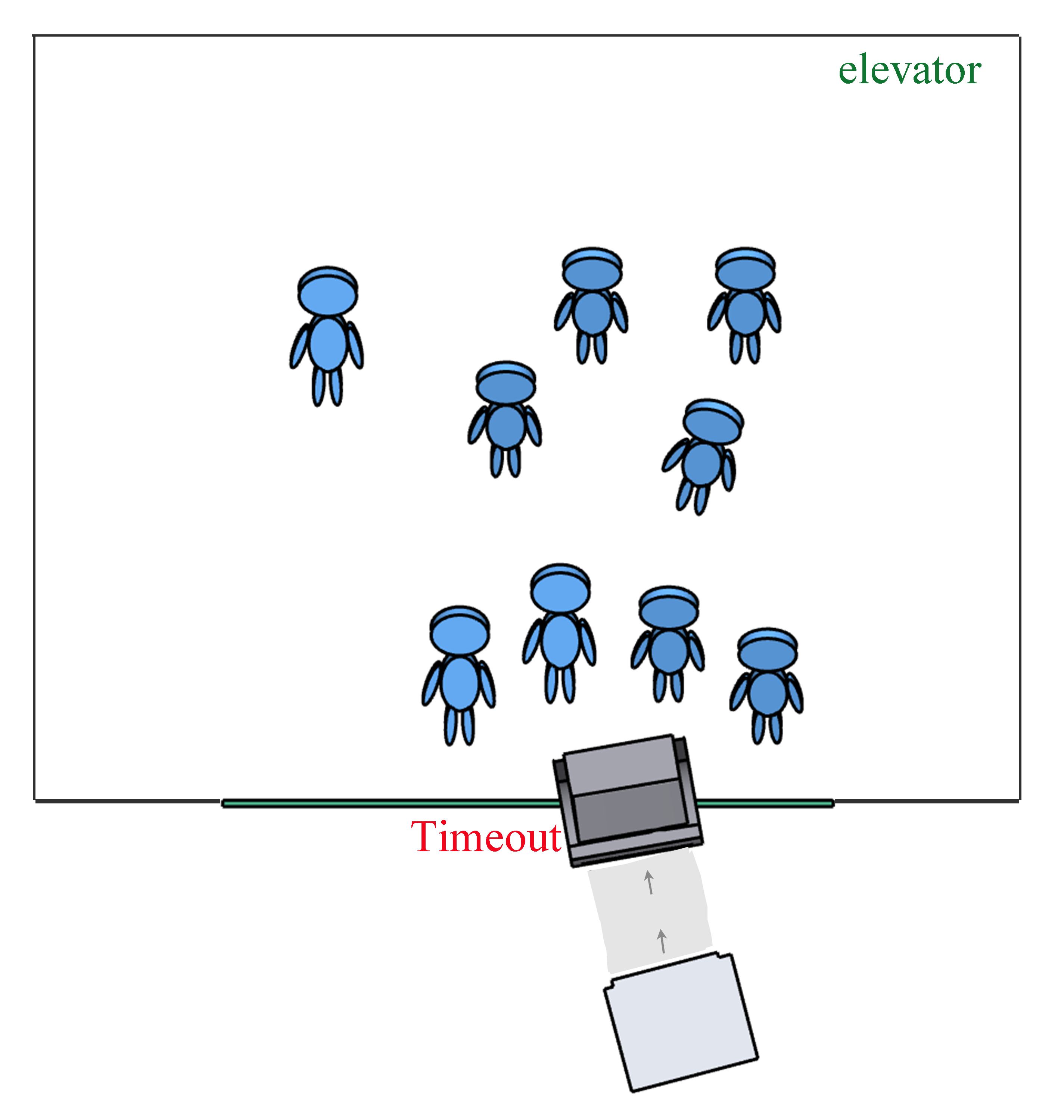}}
\quad
\subfigure[]{\includegraphics[width=3.5cm]{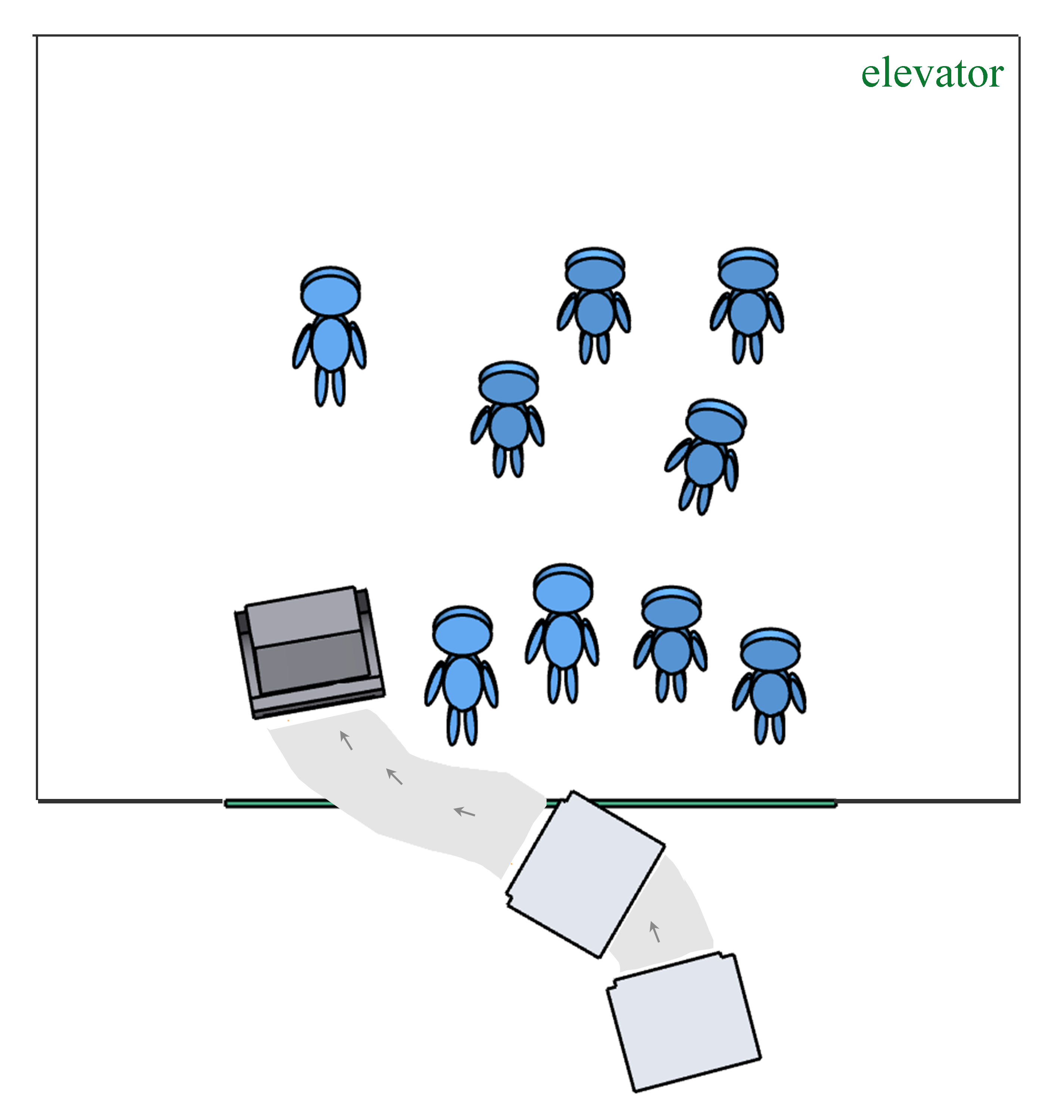}}
\quad
\subfigure[]{\includegraphics[width=3.5cm]{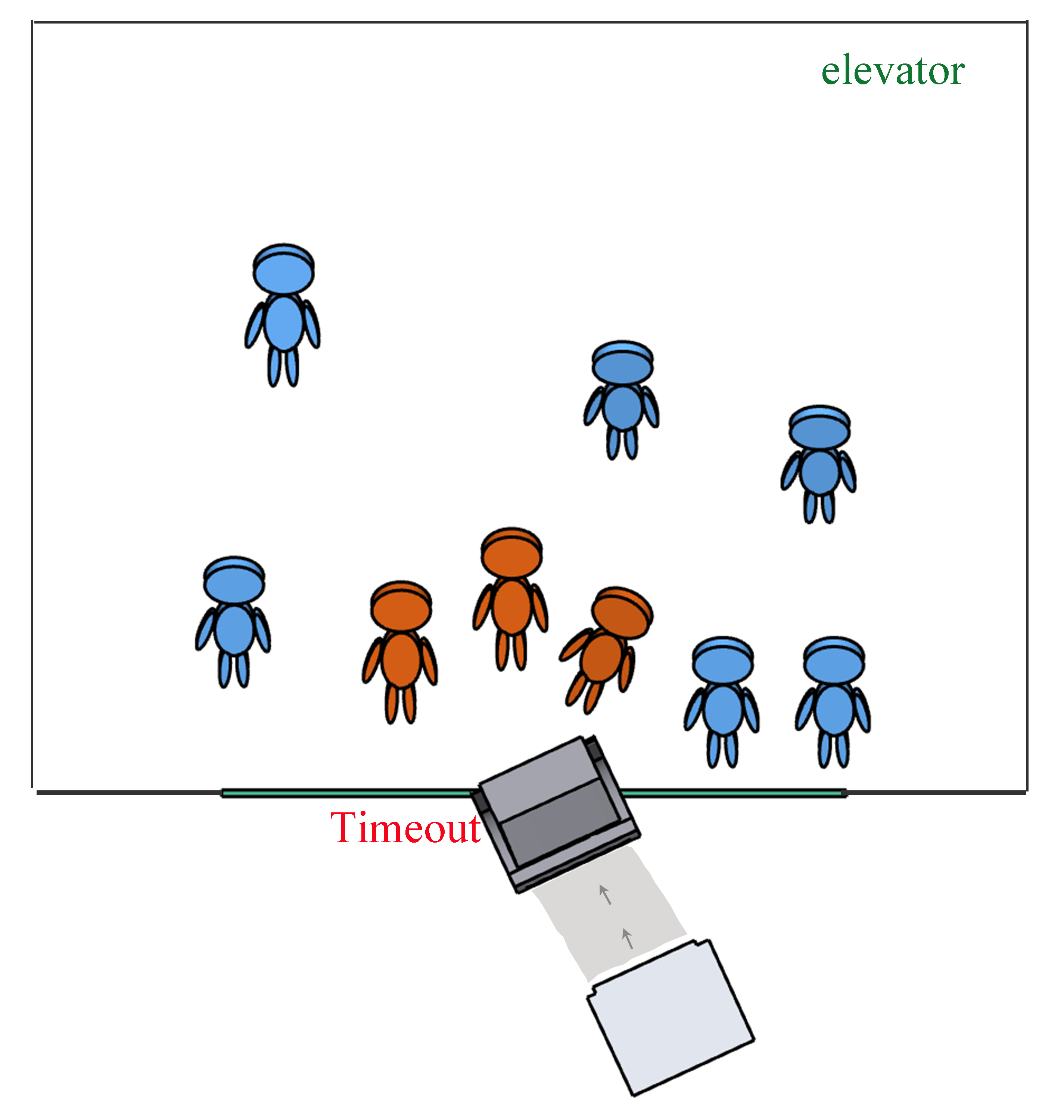}}
\quad
\subfigure[]{\includegraphics[width=3.5cm]{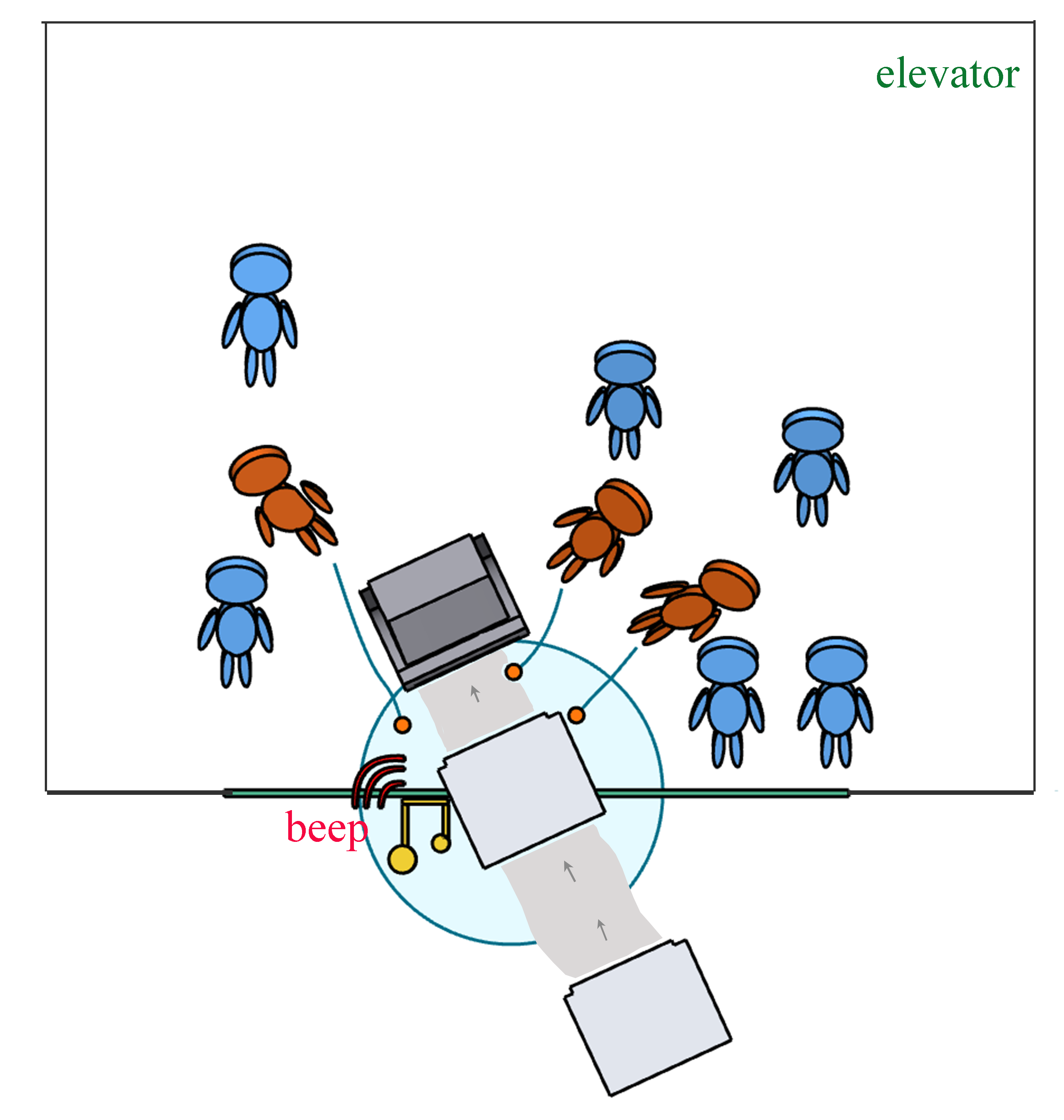}}
\caption{Demonstration for robots to enter the elevator. The initial location distribution of humans in the elevator
in (a) and (b) is the same. Similarly, (c) and (d) have the
same initial setting. (a) There is a way to the elevator, but the navigation algorithm experience a timeout, because it lacks of exploration ability; (b)The robot with high exploration ability explores a new path to go; (c) There is no available way to the elevator, and the robot is failed to enter the elevator and experience a timeout; (d) navigation policy with active clear path action successfully enters the elevator with the cooperation of people. The red-coloured human figures in (c) and (d) refer to those who block the robot's path to the elevator and require to make room for the robot.}
\label{fig:ORPR}
\end{figure}

To address this problem, we proposed a flexible and online strategy for mobile robots to enter the elevator with good consideration for safety and efficiency. The strategy is based on reinforcement learning, which calculates the optimal path and determines whether to make interaction with humans in real time. Thus, the proposed navigation policy enables the robot automatically chooses between passively avoiding obstacles (Shown in Fig. \ref{fig:ORPR} (b)) by finding another path  and actively clearing a planning path by voice interacting with humans depending on the current distribution of people and obstacles. Particularly, when there is enough room in the elevator for the robot to safely fit in but its path is blocked, the model has learned from the knowledge of the current environment to obtain the potential ability to recognize it and then try to clear the forward path to enter the elevator, (Please see Fig.~\ref{fig:ORPR} (c) and (d), The red-coloured human figures refer to those who block the robot's path to the elevator and require to make room for the robot) rather than wasting time for waiting for the next elevator.

In summary, the following are the vital contributions of this work: 
\begin{enumerate}[]
\item To the best of our knowledge, to solve the problem of robots efficiently entering the elevator, this paper is the first research outcome to combine a reinforcement learning-based navigation algorithm with voice interaction to improve the efficiency of mobile robots on the basis of safety.

\item We proposed an efficient deep neural network to acquire optimal state value in reinforcement learning. The network implements a channel-attention mechanism based on a transformer encoder to weight channels in feature level for the purpose of acquiring more accurate state value.
\end{enumerate}

The rest of the paper is organized as follows: Section II shows the backgrounds and related work on mobile robot navigation strategies whilst in Section III, the solution for entering and exiting the elevator securely and efficiently is presented. The experimental surroundings and parameter settings are described in Section IV, and the qualitative and quantitative analyses of the experimental findings are performed in Section V. Section VI concludes the paper with a summary and future works.

\section{Background \& related work}

\subsection{Safety and efficiency robot navigation}

For the purpose of making the navigation of the robot safe, various obstacle avoidance methods are proposed in the literature and industry. These navigation algorithms generally can be divided into two classes, which are traditional hand-crafted functions based and machine learning based, respectively. The traditional navigation methods \cite{koren1991potential, van2008reciprocal, van} navigate robots based on some particular presetting navigation functions, which are designed to dodge obstacles during the process of reaching the targets for the consideration of safety. The drawback of this kind of method is that they heavily depend on hand-crafted functions which are hard to fit various scenarios. Other kinds of navigation methods are based on machine learning, such as CADRL\cite{chen2017decentralized} and SARL \cite{Chen19}. These methods are typically based on reinforcement learning due to the navigation process can be regarded as a Markov decision process (MDP). Since those methods train their model by very huge datasets which are generated randomly in their reinforcement learning method, those methods have the ability to generalize to various scenarios even in moving crowds. That is also one of the reasons why RL-based agents show the potential to give a better solution compared to the human conceivable solutions in many other applications such as AlphaGo \cite{silver2016mastering} or AlphaStar \cite{arulkumaran2019alphastar}. 

Both the two kinds of navigation methods mentioned above can provide strong driving safety for low-speed robots because their aim is obstacle avoidance. Specifically, the operation of obstacle avoidance for the two kinds of methods is to go by a roundabout route. However, always making way for obstacles obviously affect the navigation time. Meanwhile, when the environment is crowded and humans blocked the routes of robots, these methods generally wait until an available way is found, which are time-consuming.

In this case,  Robot-Human interaction is required to improve the efficiency of navigation. Human-computer interaction \cite{3,4,5,6,7} has been developed for many years for the purpose of increasing the efficiency of robots, user experience, etc. Among them, voice interaction \cite{yang2020hybrid, matsusaka2009health, khalid2016exploring, anzai1993human} is the most popular mode. The changes for mobile robots to actively interact with humans is how to find a suitable time and location to execute the interaction. To answer this question, we investigate some robots in the industry such as Yunji and Keenon and some methods in the literature. Most of them use some presetting judgment conditions, and when those conditions are triggered the robots will take voice interaction. As presetting judgment conditions cannot be generalized to various scenarios, more flexible and intelligent interaction methods are required to improve the efficiency of mobile robots.

\subsection{Robot navigation in elevator environment}

Compared to large-area navigation tasks, the elevator environment is smaller and generally more crowded. This situation requires that the navigation algorithm has a stronger ability to explore feasible routes based on the limited information of the current environment. Meanwhile level of navigation aggressiveness is required higher than that in other open places. This is because no matter how the robot avoids obstacles, it will not relieve the congestion of the elevator but waste time for the next elevator. However existing navigation approaches \cite{34,38,35,36} in the literature mainly focus on the safety of mobile robots instead of efficiency. As we know, in muli-floor tasks, waiting for an elevator occupies a large proportion of the entire shipping time for delivery robots. Thus, the efficiency of entering the elevator does affect the efficiency of the entire shipping task. However, the traditional navigation algorithms \cite{37,11,12,13,29} might be not helpful in that crowded environment because of their limited exploring route ability and they are highly likely to lead the robot to freeze~\cite{31} in the real-world application. A flexible navigation policy and an efficiently active clear path operation might be a possible solution for solving that problem. Considering reinforcement learning based navigation methods \cite{19, 20, 32} shows outstanding exploring ability for various unknown environments and the efficiency of voice interaction, we proposed our approach based on these two technologies.


\section{Approach}

\subsection{Reinforcement Learning for Robot Navigation}

A robot navigation task of entering the elevator can be formulated as an MDP, which can be solved by A reinforcement learning framework \cite{Nishimura2020L2BLT}. The MDP can be defined by a tuple $M = \left\{S,A,P,R,\gamma\right\}$  \cite{everett2021collision}. 
\begin{itemize}
    \item $S=\left\{s_{1}, s_{2}, \ldots, s_{n}\right\}$ is the set of states. 
    \item $A=\left\{a_{1}, a_{2}, \ldots, a_{m}\right\}$ is the set of actions.
    \item $P: S \times A \mapsto S$ are the state transition probabilities. For each state $s \in S$ and action $a \in A$, $T_{a}\left(s, s^{\prime}\right)=\mathbb{P}\left[S_{t+1}=s^{\prime} \mid S_{t}=s, A_{t}=a\right]$ denotes the probability of reaching state $s^{\prime} \in S$.
    \item $R: S \times A \mapsto \mathbb{R}$ are the reward function. For each state $s \in S$ and action $a \in A$, $R_{a}(s,s^{\prime}) = r$ denotes the reward for reaching state $s^{\prime} \in S$.
    \item $\gamma \in[0,1]$ is the discount factor.
\end{itemize}

Specifically, The state $s^{\mathrm{R}}$ for robot is defined by

\begin{equation}
	s^{\mathrm{R}}=\left[d_{\mathrm{g}}, \boldsymbol{v}_{x},\boldsymbol{v}_{y}, v_{\text {pref }}, r \right]
\end{equation}
where $d_{g}=\left|p_{y} - p_{y}^{\mathrm{g}}\right|$ is the vertical distance from the current position of the robot to the interior of the elevator. The reason behind considering only the vertical distance is that the aim of the robot is to enter the elevator rather than reach a particular location point. The vector $\mathbf{v}=\left[v_{x}, v_{y}\right]$ is velocity whilst $\mathbf{p}=\left[p_{x}, p_{y}\right]$ is the current position. Both preferred speed $v_{\text {pref}}$ and the robot radius $r$ are hyper-parameters.

Similarly, the state of the $i$ th human is given by
\begin{equation}
	s_{i}^{\mathrm{H}}=\left[p_{x}, p_{y}, v_{x}, v_{y}, r_{i}, d_{i}, r_{i}+r\right]
\end{equation}
where $d_{i}=\left\|\mathbf{p}-\mathbf{p}_{g}\right\|_{2}$ presents the distance between the robot and the $i$ th human; $r_{i}$ is radius of the $i$ th human.

Thus, the joint state of the RL framework becomes 
\begin{equation}
	s^{\mathrm{jn}}=\left[s^{\mathrm{R}}, s^{\mathrm{H}}\right]
\end{equation}

Regarding the robot movements, we set 17 actions in the action space of the robot (independent from beeping), which consist of 16 orientations evenly spaced between [0,2$\pi$) at the velocity $v_{p r e f}$ and a stay action. ~On the other hand, when considering clear path action, the number of actions in action space is doubled to 34 since each moving action is added with a sound warning (beep).

Assuming the robot and humans are controlled separately by their own navigation policies in the simulation, the humans in the elevator can, therefore, be regarded as a part of the environment. We adopt the state transition probabilities $P\left(\mathbf{s}_{t+1}^{j n} \mid \mathbf{s}_{t}^{j n}, \mathbf{a}_{t}\right)$ based on Optimal Reciprocal Collision Avoidance (ORCA) policy \cite{van, Xu21} which controls human behaviors and some hidden parameters (velocities, goals, etc.). 

The aim of using the reinforcement learning method is to train an optimal policy $\pi^{*}$ for a robot to choose the most suitable action $a_{t}$ in every time step so as to maximize the expectation of the return of an episode, which, in the application, leads to robot entering the elevator safely and efficiently. The optimal policy and state value function in time $t$ can be defined as:

\noindent
\begin{dmath}
	\pi^{*}\left(\mathbf{s}_{t}^{j n}\right)= \underset{\mathbf{a}_{t}}{\arg\max}\left[ R\left(\mathbf{s}_{t}^{j n}, \mathbf{a}_{t}\right) + \gamma^{\Delta t \cdot v_{pref}}\times\\\int_{\mathbf{s}_{t+\Delta t}^{j n}} P\left( \mathbf{s}_{t+\Delta t}^{j n}| \mathbf{a}_{\mathbf{t}}, \mathbf{s}_{t}^{j n}\right) V^{*}\left(\mathbf{s}_{t+\Delta t}^{j n}\right) d \mathbf{s}_{t+\Delta t}^{j n}\right]
\end{dmath}
where the discount factor is $\gamma \in(0,1)$, and the probability of transition is $P\left( \mathbf{s}_{t+\Delta t}^{j n}\mid \mathbf{a}_{\mathbf{t}}, \mathbf{s}_{t}^{j n}\right) \in(0,1) $ which presents the probability of transforming to the state $\mathbf{s}_{t+\Delta t}^{j n}$ from the state $\mathbf{s}_{t}^{j n}$ by taking action $\mathbf{a}_{\mathbf{t}}$, and 
\begin{equation}\label{equ:optVal}
	V^{*}\left(\mathbf{s}_{t}^{j n}\right)=\sum_{t^{\prime}=t}^{T} \gamma^{t^{\prime} \cdot v_{p r e f}} R_{t^{\prime}}\left(\mathbf{s}_{t^{\prime}}^{jn}, \pi^{*}\left(\mathbf{s}_{t^{\prime}}^{j n}\right)\right)
\end{equation}

\begin{figure*}[t]
	\centering
	\includegraphics[width=\linewidth]{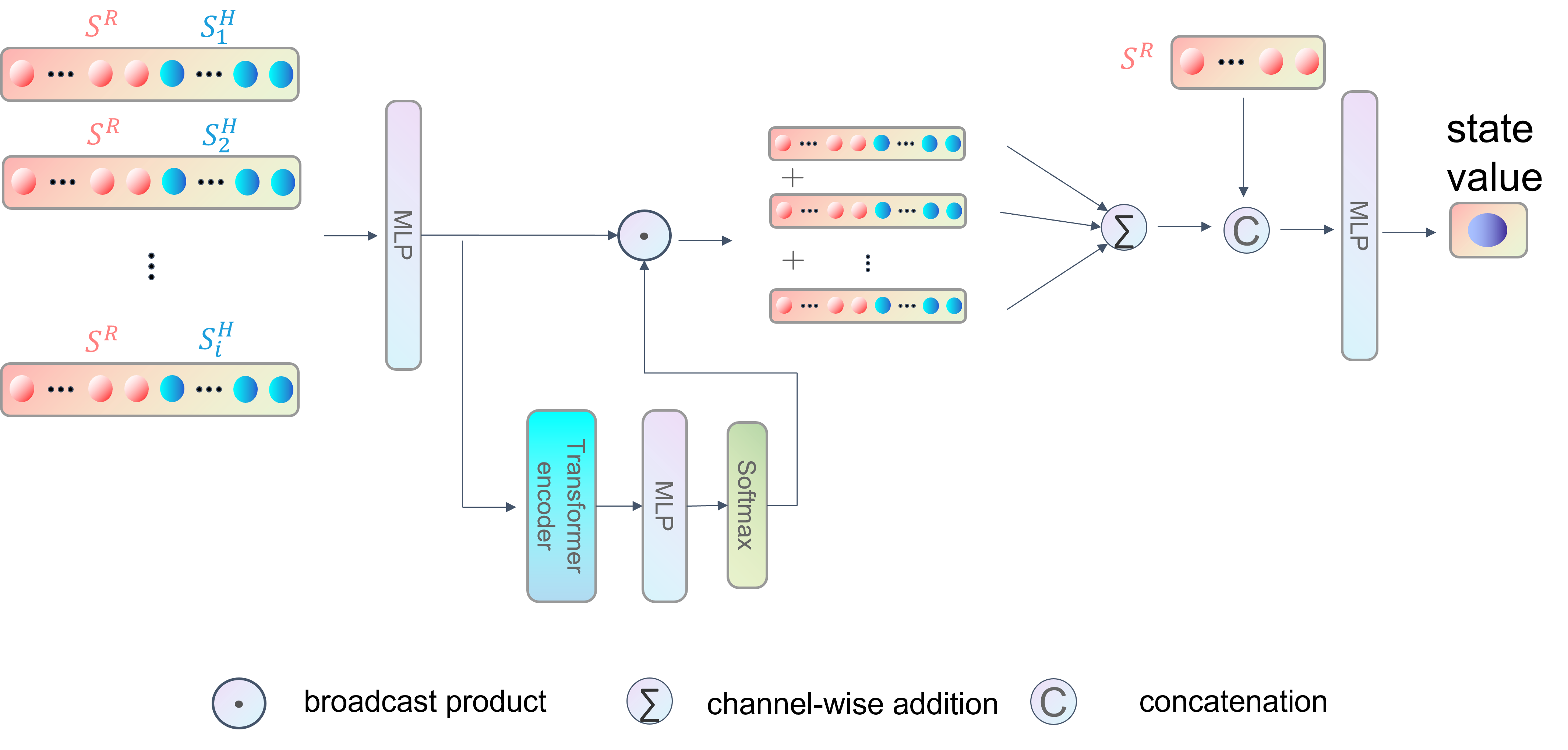}
	\caption{The NavFormer network for calculating state value. The $s^{\mathrm{R}}$ and $s_{i}^{\mathrm{H}}$ refer to states of robot and humans respectively.}
	\label{fig:netStruct}
\end{figure*}

In order to obtain the optimal value function $V^{*}$ in (\ref{equ:optVal}) efficiently, we proposed a transformer based deep neural network, which is named as NavFormer and shown in Fig.~\ref{fig:netStruct}. The combination of joint states for all agents in the environment is input into a multilayer perceptron (MLP). A transformer encoder \cite{vaswani2017attention} is implemented to calculate the weights for each channel of the high-level future. The weights are multiplied by the output of MLP to make it more informative. Then, each channel of weighted features is added together. In order to make use of low-level features, the robot state is concatenated with the sum of weighted features before the final MLP to output the state value.

\subsection{Learning to Balance Between Safety And Efficiency}

When a robot is navigated in an aggressive way (high speed, high-frequency clear path, etc.), even though it generally saves time, this is highly likely to affect safety. The proposed strategy adopted the safety-based navigation algorithm but furthermore adds efficiency consideration. Namely, safety is also an important factor in our approach whilst we designed the method to be more efficient on the basis of safety. Safety-based navigation algorithms may waste time in the elevator environment. For instance, there is a highly common human behavior phenomenon that people stand in front of the elevator door and block the entrance even though there is still a lot of room inside the elevator. In cases like this, most of the existing robot navigation policies suggest a time-consuming action to wait for the next elevator for safety considerations. An active clear path action can be an efficient method to overcome this problem and improve the efficiency of robots to complete tasks in various elevator scenarios. In the simulation, it is easy to acquire the locations of each people in the elevator. Meanwhile, when the active voice interaction happened, people will make room for the robot based on Reciprocal Velocity Obstacles (RVO) \cite{van2008reciprocal} algorithm, which can be used to imitate humans to dodge obstacles according to its design. Also, in real-world experiments, a 360$\degree$ lidar scanner has the potential to recognize this situation by measuring the elevator space through the gap between the human legs, which provides the feasibility for improvement on efficiency for mobile robots in elevator environments. A speaker is used for voice interaction, which is described in Section IV in detail.

To realize the idea above, we designed a reward function in reinforcement learning. On the one hand, Following \cite{everett2021collision} a part of reward function $R_{t}^{safety}\left(\mathbf{s}_{t}^{j n}, \mathbf{a}_{t}\right)$ is designed to guarantee safety, which penalizes collisions and uncomfortable distance between the robot and humans. Meanwhile, in order for the robot to complete the task as quickly as possible, the reward is also designed to be related to the task completion time. Every moving step of the robot needs to pay a small price for the purpose of finding the shortest path. Thus, the reward function for safety  $R_{t}^{safety}$ is defined by

\begin{equation}\label{equ:reward}
	R_{t}^{safety}\left(\mathbf{s}_{t}^{j n}, \mathbf{a}_{t}\right) = \left\{\begin{array}{ll}
		-0.25                       & \text { if } d_{t} < 0 \\
            -2                          & \text { else if } t > T \\
		10+\left(T-t\right)*0.15    & \text { else if } \mathbf{p}_{t} = \mathbf{p}_{g} \\
		0                           & \text { else if } v = 0  \\
            -0.1+d_{t} / 2              & \text { else if } d_{t} < 0.2 \\
		-0.01                       & \text { otherwise }\\
	\end{array}\right.
\end{equation}

where $ d_{t} $ refers to the closest distance between the robot and humans at time $t$. $d_{t} < 0$ means there is a collision between them. In order to avoid a collision, we set the reward of collision as a negative value of -0.25 to penalize it. $T$ represents the time limit to complete the task. $t>T$ means the task is timeout, which is penalized by a negative reward value of -2. At the time $t$, $v$ and $\mathbf{p}_{t}$ are the current velocity and the position of the robot, respectively. When $\mathbf{p}_{t}$ equals to $ \mathbf{p}_{g}$, the robot enters the elevator. In this case, the agent is awarded a positive reward, which is also related to the finish time. The shorter the finish time, the bigger the reward. If none of the above conditions is triggered, the robot still needs to pay a small price of -0.01. The robot is expected to be stopped ($v = 0$ no need to pay a price) rather than run somewhere when there is no available path to the elevator for the robot to go. Since the judgment conditions cannot form a group of collectively exhaustive events, these conditions are executed sequentially from top to bottom.

On the other hand, to consider the efficiency of the robot, a reward for actively clear path action is added. When people stand in front of the elevator door and the robot realizes there is still usable room in the elevator, it could take a beep action to make humans make room for it. However, beeping too frequently may cause too much noise and not smart enough. In order to make the robot's behaviour more natural and avoid the abuse of voice interaction, every beep action needs to pay a price. The final reward function is given by

\begin{equation}\label{equ:reward2}
R_{t}\left(\mathbf{s}_{t}^{j n}, \mathbf{a}_{t}\right) = \left\{\begin{array}{ll} 
 R_{t}^{safety}\left(\mathbf{s}_{t}^{j n}, \mathbf{a}_{t}\right) - 0.1 & \text { if beep } \\
 R_{t}^{safety}\left(\mathbf{s}_{t}^{j n}, \mathbf{a}_{t}\right)  & \text { otherwise } \\
 \end{array}\right.
\end{equation}

\section{Experiments}
\subsection{Simulation Setup}
In the simulations, we followed a unitless approach so that they can be rescaled to fit the real environments. The crowd in the elevator was simulated by circles with radius 1 in an 8$\times$12 rectangle. Similarly, the robot was also regarded as a circle with a radius of 1 but filled with a particular colour different from humans. The locations of the crowd were randomly distributed inside the rectangle in each training iteration. In the simulation, when the robot made a clear path action (Beep), the RVO navigation algorithm is used to imitate people making room for the robot without collision. Here, we set the safety space between humans as 0.2, and the max velocity for humans as 1. According to the requirement of the RVO navigation algorithm which is used for navigating humans to make room, each agent (human) needs to be given a destination and a starting point. We initial the destination of each human as its starting point to imitate the general circumstance that people stand in the elevator. Once the robot takes an active clear path operation, according to the RVO algorithm, the human will move along the vertical direction of the robot's velocity to make room for the robot. Specifically, the moving velocity and moving distance of humans are related to the moving direction of the robot and the relative position between the robot and the human. The used simulation platform for imitating humans is the Python lib of RVO.

For the robot navigation policy, the starting point is set on the outside of the elevator facing the middle of the elevator door and the distance between them is 3. The destination of the robot is anywhere inside the elevator which is realized by measuring the vertical distance between the robot and the door. The preferred speed $v_{p r e f}$ of the robot is set to 1. The task will end only when the robot
\begin{enumerate}[]
	\item enters the elevator,
	\item collides with humans,
	\item timeout (9s)
\end{enumerate}

\subsection{Training Setup}

We implemented our methods by using the Pytorch \cite{NEURIPS2019_9015} environment. Following \cite{Chen19}, Before reinforcement learning, in order to shorten training time, there is imitation learning for the robot to learn a pre-trained model based on 3,000 demonstrations from ORCA. The learning rate for imitation learning is 0.01, and the 3,000 demonstrations of ORCA are used for training as 50 epochs. For the last 7000 episodes, deep reinforcement learning is performed with a learning rate of 0.001 and a discount factor $\gamma$ of 0.9. Also, in order to randomly explore more actions in the initial training process, we set the bigger exploration rate of 0.5 for the first 5,000 episodes and then keep 0.1 for the remaining 5,000 episodes to speed up model optimization. A mini-batch stochastic gradient descent (SGD) optimizer is used in both imitation and reinforcement learning with a batch size of 100. The simulation experiments were performed by The Intel® Core™ i7-11700 processor. The real experiments are performed in the embedded AI platforms of Nvidia Jetson AGX Xavier.

\section{Experimental Results and Analysis}

\subsection{Measurement Metrics Explanation}

We analyzed all the navigation policies by 1000 random cases via related measures including the success rate of entering the elevator, collision rate, timeout rate, average navigation time for those success cases, and average total reward. The sum of success, collision, and timeout rates is 1. 

The collision rate reflects the driving safety of the robot, while the timeout rate and average navigation time refer to efficiency. In the reinforcement learning approach, the agent will get a reward for each step to reward or punish every action. The sum of rewards during one episode reflects the overall performance of this episode, which is called \textit{return}. In the evaluation stages, we calculate the average return of 1000 episodes, which means the average performance of 1000 cases can be obtained. The reward is defined by the reward function, which is considered to guarantee driving safety and improve efficiency. Thus, the average total reward can be regarded as a comprehensive evaluation. When the average total reward value is higher, that means the navigation algorithm works more consistently with the expectation which is defined by the reward function. 

\subsection{Quantitative Evaluation}

\begin{table*}[t]\footnotesize
	\centering
	\caption{Quantitative evaluation of efficiency for each navigation algorithm} \label{Table:1}
	\setlength{\tabcolsep}{1mm}{
	\begin{tabular}{c|c|c|c|c|c|c|c|c|c}
        \toprule[0.75pt]
                               &    \multicolumn{3}{c|}{6 people inside elevator}& \multicolumn{3}{c|}{7 people inside elevator} &  \multicolumn{3}{c}{8 people inside elevator} \\
        \cline{2-10} 
             Method            & Success     &  Timeout    &  Time (s)   & Success     &  Timeout     & Time (s)  & Success   &  Timeout      &  Time (s) \\
        \midrule[0.5pt]
        ORCA \cite{Xu21}       &   0.58      &  0.42       &    6.24     &   0.37      &   0.63      &   6.78      & 0.26         &  0.74        & 6.98   \\  
        SARL \cite{Chen19}     &   0.84      &  0.14       &    4.78     &   0.66      &   0.30      &   4.95      & 0.69         &  0.26        & \textbf{4.93}   \\
        NavFormer (w/ Beep)    &\textbf{0.99}&\textbf{0.00}&\textbf{4.64}&\textbf{1.00}&\textbf{0.00}&\textbf{4.63}& \textbf{0.84}& \textbf{0.14}& 5.30   \\ 
        \bottomrule[0.75pt]
	\end{tabular}}
\end{table*}

\begin{table*}[h]\footnotesize
	\centering
	\caption{Quantitative evaluation of Collision rate and Total Reward for each navigation algorithm} \label{Table:2}
	\setlength{\tabcolsep}{1mm}{
	\begin{tabular}{c|c|c|c|c|c|c}
        \toprule[0.75pt]
                              &    \multicolumn{2}{c|}{6 people inside elevator}& \multicolumn{2}{c|}{7 people inside elevator} &  \multicolumn{2}{c}{8 people inside elevator} \\
        \cline{2-7} 
             Method            & Collision &   Reward      &  Collision &    Reward   & Collision &  Reward  \\
        \midrule[0.5pt]
        ORCA  \cite{Xu21}      &\textbf{0.00}&    2.9195     &\textbf{0.00}&     1.4607    &\textbf{0.00}&      0.7264        \\  
        SARL \cite{Chen19}     &   0.02      &    5.5359     &   0.04      &     4.6234    &    0.05     &      4.3443        \\
        NavFormer (w/ Beep)    &   0.01      &\textbf{6.7040}&\textbf{0.00}&\textbf{6.6787}&    0.02     &  \textbf{4.9035}   \\ 
        \bottomrule[0.75pt]
	\end{tabular}}
\end{table*}


We conducted three groups of control experiments to test the performance of each model according to corresponding measure indexes. Three groups of control experiments are ORCA, SARL, and the proposed NavFormer. In the simulation, only the proposed method used a voice interaction, because it is hard to find a suitable moment to take a clear path for ORCA. If a robot always clears its path, it obviously too aggressive and abnormal. In each group of control experiments, the number of people in the elevator varies between 5 and 8. Table \ref{Table:1} shows the performance of efficiency, which involves success rate, timeout rate, and average navigation time. The success rate and timeout rate get raised with the increase in the number of people inside the elevator. This is because,  when the number of humans grows, the number of available paths to the elevator decrease, and the probability of people blocking the door of the elevator rises. However, the ability to explore new paths and the range of obstacle avoidance of ORCA policy are limited. When their route is blocked, they choose to wait for the next elevator, which causes high timeout rates. On the other hand, the RL-based SARL navigation method has a stronger ability to explore new paths to the elevator, and the success rate is much higher than that of ORCA among the 3 experiment groups. Furthermore, The proposed method increases the performance again. Especially, when the number of people is 8, the success rate reaches 84\% and the timeout rate is only 14\%, which are much better than other methods. 

Another efficiency index is the average navigation time for those cases in which the robot enters the elevator without collision. The proposed method shows the best performance when the number of people is 6 or 7, but SARL exceeds NavFormer and reaches the best for the cases with 8 people inside the elevator. We consider the reason is that time is consumed when the humans make room for the robot. Even so, the proposed method reduced the timeout rate by taking voice interaction, which still improves overall efficiency. It is worth noting that the success rate of SARL decreases from 84\% to 69\% as the number of people increases from 6 to 8. This huge success degradation is not surprising since the RL without beep policy cannot address the situation where humans stand in front of the elevator door and block the way into the elevator. However, the proposed method with beep shows a great improvement in success rate with up to 100\% in the case of 7 people in the elevator.

The collision rate is regarded as the safety evaluation index. Since ORCA is designed to avoid collisions, it has the best performance for safety, but the collision rates of the other two methods are also very low. Thus, safety for all models is very similar, but the efficiency has huge differences. As SARL and the proposed method use the same reward function in terms of safety, their collision rates are very similar but NavFormer is slightly better because of the more efficient design of the value network based on the transformer encoder. When it comes to the average total reward, NavFormer is much better than other methods in all experiment groups. The high reward results show that NavFormer is more in line with our expectations, which shows good efficiency on the basis of safety.

\subsection{Qualitative Evaluation}

\begin{figure}[t]
	\centering
	\setlength{\abovecaptionskip}{0.cm}
	\setlength{\belowcaptionskip}{-0.cm}
	\setlength{\dbltextfloatsep}{0pt}
	\subfigure[]{
		\includegraphics[width=0.45\linewidth]{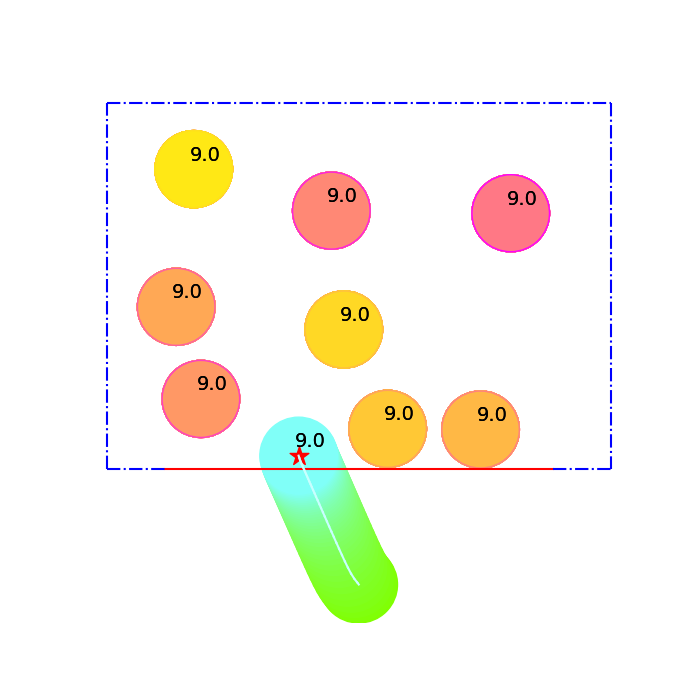}
	}
	\subfigure[]{
		\includegraphics[width=0.45\linewidth]{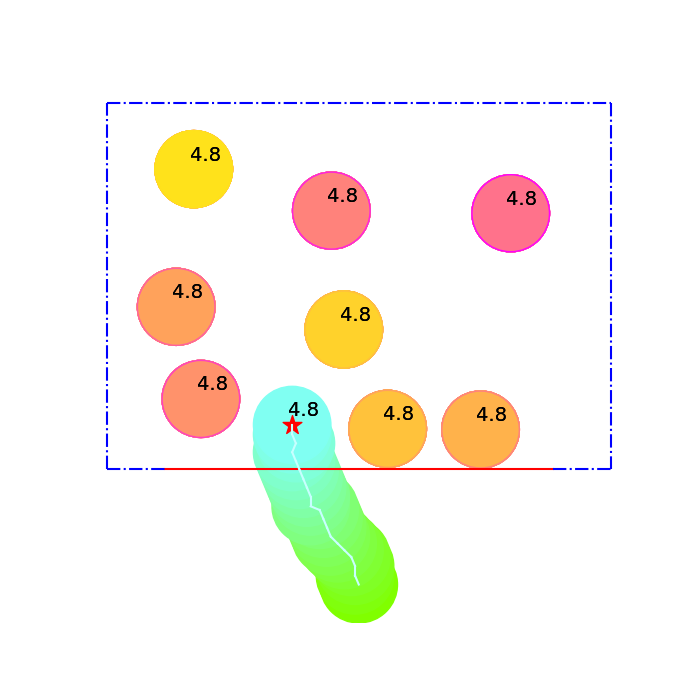}
	}
	\subfigure[]{
		\includegraphics[width=0.45\linewidth]{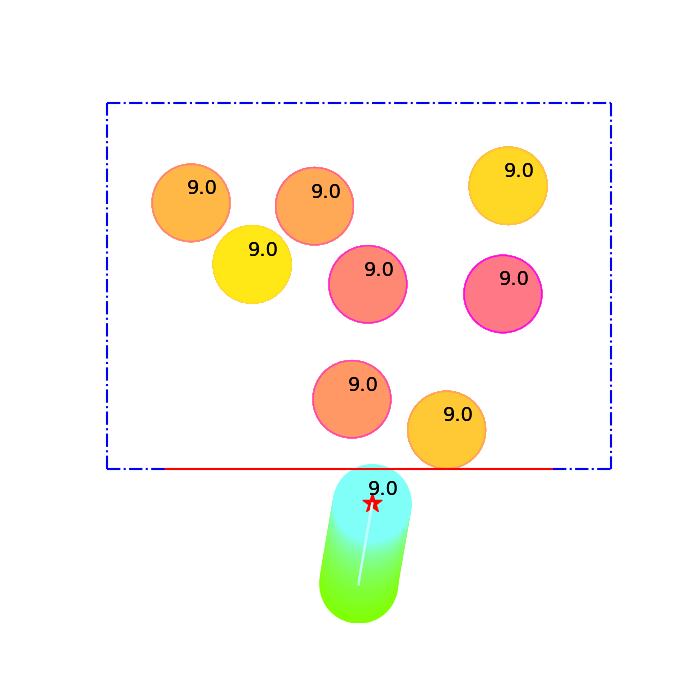}
	}
	\subfigure[]{
		\includegraphics[width=0.45\linewidth]{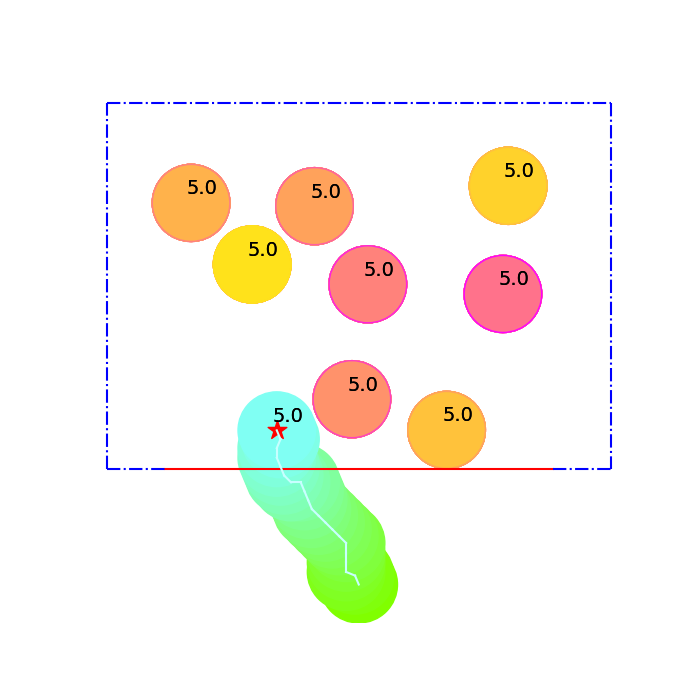}
	}
	\caption{Performance evaluation for a no-beep scenario. (a) (c) Cases for ORCA policy; (b) (d) The same cases in (a) and (c) but for RL policy.}
	\label{fig:noBeep}
\end{figure}

There are some cases shown for qualitative analysis. In Fig.~\ref{fig:noBeep}, the initial location distribution of humans in the elevator in (a) and (b) is the same. Similarly, (c) and (d) have the same initial setting. In both cases, there are enough spaces for a robot to enter the elevator. However, a traditional ORCA-based robot fails to enter the elevator with a timeout (9s) in both cases (a) and (c). This is because the objective of ORCA based method is to reach a goal according to its initial setting. Meanwhile, it is hard for this kind of traditional navigation algorithm to change its path in real-time according to the environment. As a result, the method is more likely to waste time waiting for the next elevator and loss of efficiency for the delivery robots. 

On the contrary, RL based method shows an advantage of path re-planning, which has a certain ability to explore other feasible routes when its current path is blocked. As shown in Fig.~\ref{fig:noBeep}, both (b) and (d) experience success in the task of entering the elevator which cost 4.8 and 5 seconds respectively. RL policy enables the robot to choose the optimal action in each time step (0.25s) according to the current environment instead of the initial path setting. Thus, RL based navigation method shows more flexible and effective performance in the task of entering the elevator than the traditional navigation algorithms such as ORCA. Voice interaction does not execute in both of the cases above, because in these scenarios it is not necessary to clear the path for entering the elevator.

\begin{figure}[t]
	\centering
	\setlength{\abovecaptionskip}{0.cm}
	\setlength{\belowcaptionskip}{-0.cm}
	\setlength{\dbltextfloatsep}{0pt}
	\subfigure[]{
		\includegraphics[width=0.45\linewidth]{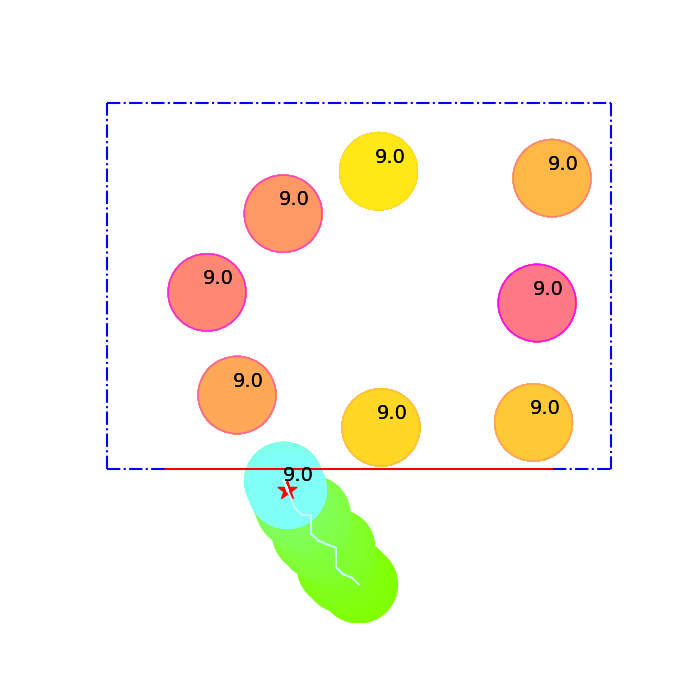}
	}
	\subfigure[]{
		\includegraphics[width=0.45\linewidth]{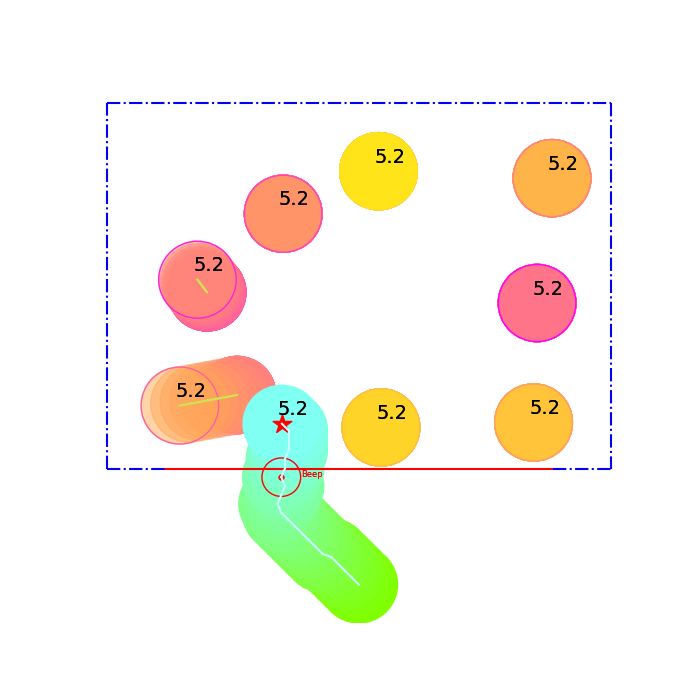}
	}
	\subfigure[]{
		\includegraphics[width=0.45\linewidth]{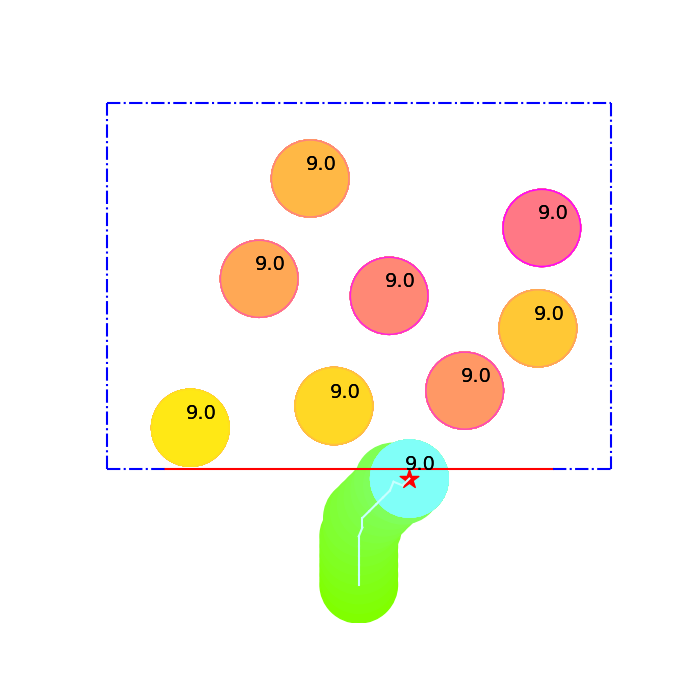}
	}
	\subfigure[]{
		\includegraphics[width=0.45\linewidth]{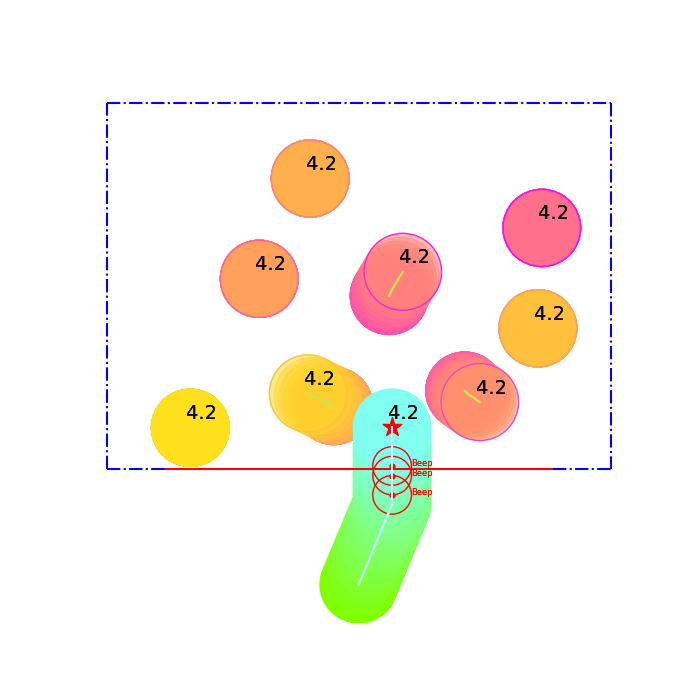}
	}
	\caption{Voice interaction (beep) performance analysis.  Cases (a) (c) for RL policy without beep; (b) (d) RL policy with a beep. Red circles in the way of the robot show positions where the robot beeps.}
	\label{fig:beep}
\end{figure}

On the other hand, in many cases, voice interaction (beep) between the robot and humans makes a huge difference in the efficiency of entering the elevator. As the two cases shown in Fig.~\ref{fig:beep}, (a),(b) and (c),(d) have the same initial settings of human locations, respectively. In these two cases, obviously, there is no available way for the robot to enter the elevator even though enough space is available within the elevator. People need to make room for the robot in order to enable the robot to enter the elevator safely. In this case, an active voice interaction becomes very important to clear the path. Cases (a) and (c) show the results of our RL navigation policy without beeping. Although the robot can flexibly replan routes, it still cannot find a feasible way to enter the elevator. However, when considering beep action (clear path) in RL policy, it changes the results. As expected, the robot executes clear path operations at suitable times and locations, which are shown as the small-red circles inside the path of the robot. After that, humans who block the entrance move to give the robot a valid route to enter the elevator. Finally, the robot enters the elevator successfully without collision. It is worth noting that since the robot needs to pay a price for each clear path action (beep), the robot executes a beep operation only when it is necessary.

\subsection{Real Experiment Result Analysis}
\begin{figure}[htbp]
	\centering
	\includegraphics[width=0.75\linewidth]{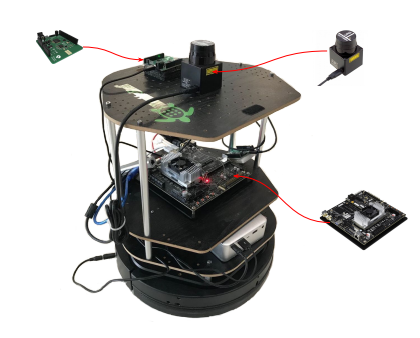}
	\caption{The Turtlebot platform}
	\label{fig:robot}
\end{figure}

In order to acquire the performance of the proposed approach in the real-world application, we deploy our reinforcement learning local planner to a Turtlebot-based platform and conduct various tests in different elevator scenarios. The hardware setup and environmental parameters are shown as follows:

\textbf{(1)Robot}: The mobile robot uses the 360$\degree$ laser scanner of RPLDAR A3 to sense the environment. An SPK speaker is implemented for the robot to execute the active voice interaction, which is beeps plus some words (the Chinese language in this work). An onboard computer of Nvidia Jetson AGX Xavier is used as the central processor. The mobile platform of the robot is Turtlebot 3, which is controlled via Revenue Online Service (ROS). The picture of the robot is shown in  Fig. \ref{fig:robot}.

\begin{figure}[htbp]
	\centering
	\includegraphics[width=0.75\linewidth]{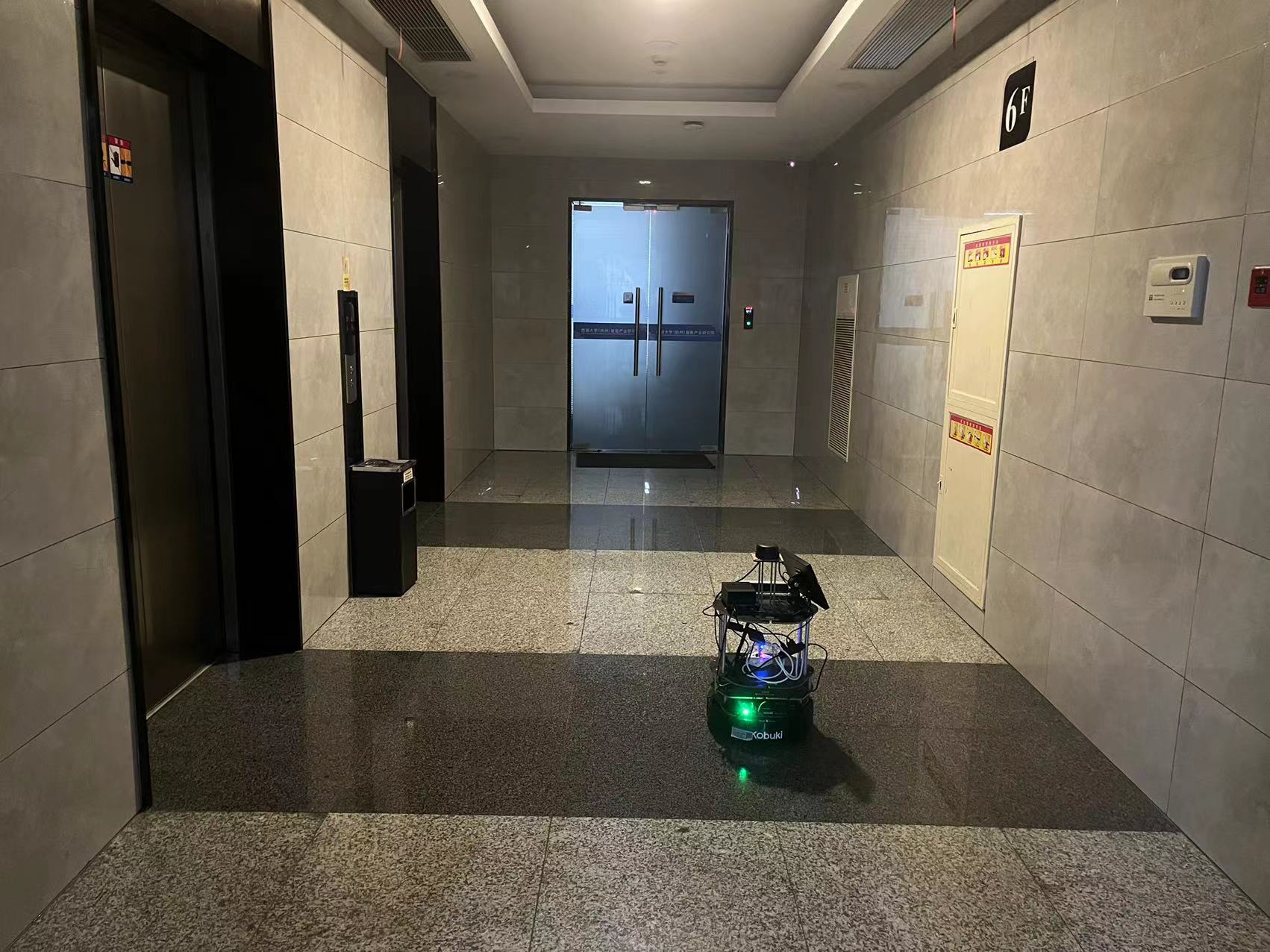}
	\caption{The experimental environment}
	\label{fig:experimental_environment}
\end{figure}

\begin{figure}[htbp]
	\centering
	\includegraphics[width=0.75\linewidth]{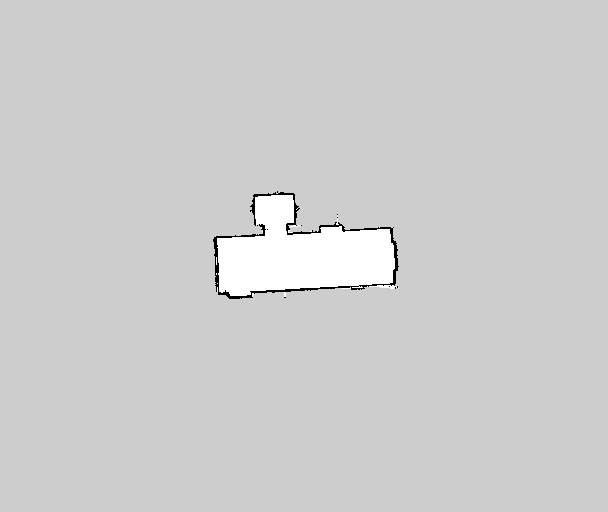}
	\caption{Environment map of real world}
	\label{fig:real_world_map}
\end{figure}

\textbf{(2)Environmental parameter}: The elevator is 1.5 meters wide and 2 meters long, and the environment showed in \ref{fig:experimental_environment}. In order to conveniently apply the proposed method in the robot system, based on the experimental elevator environment, we calculate the locations of humans and establish the map shown in \ref{fig:real_world_map} by using gmapping\cite{basavanna2020mapping} on ROS, which is a laser-based SLAM (Simultaneous Localization and Mapping) approach.

There are two cases shown in this subsection for real experiment analysis. 

\begin{figure}[htbp]
	\centering
	\includegraphics[width=0.75\linewidth]{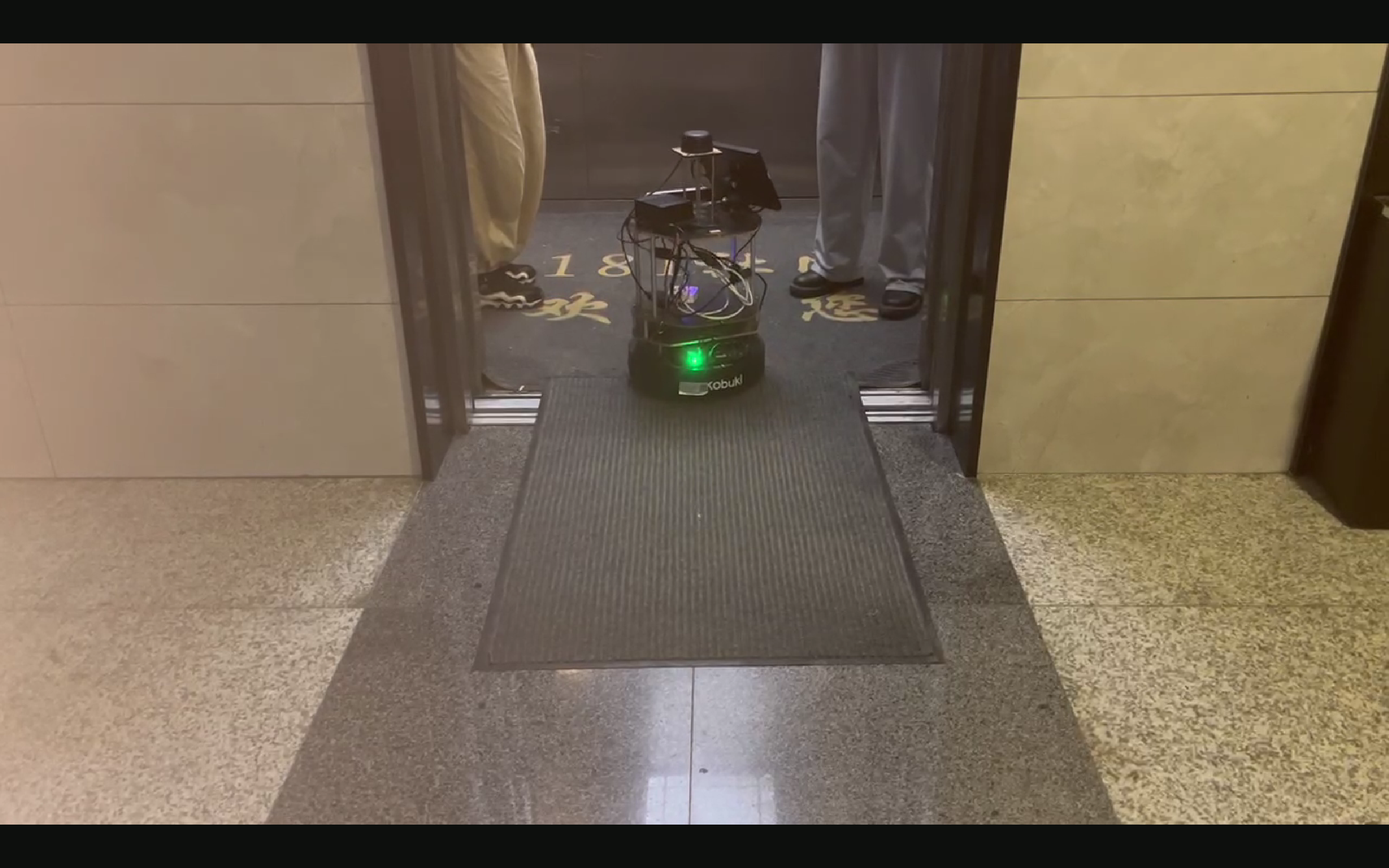}
	\caption{Humans are far away from the elevator door. The robot enters the elevator without voice interaction.}
	\label{fig:nobeep}
\end{figure}

\textbf{(1)Humans are far away from the elevator door}: In this case, it's plenty of room left for the robot to enter the elevator, and the current path of the robot is not blocked. The simulation results show that the robot could enter the elevator with no beep. As expected, the real experiment results also show that the robot could enter the elevator with no beep, which shows in Fig. \ref{fig:nobeep}.

\begin{figure}[htbp]
	\centering
	\includegraphics[width=0.75\linewidth]{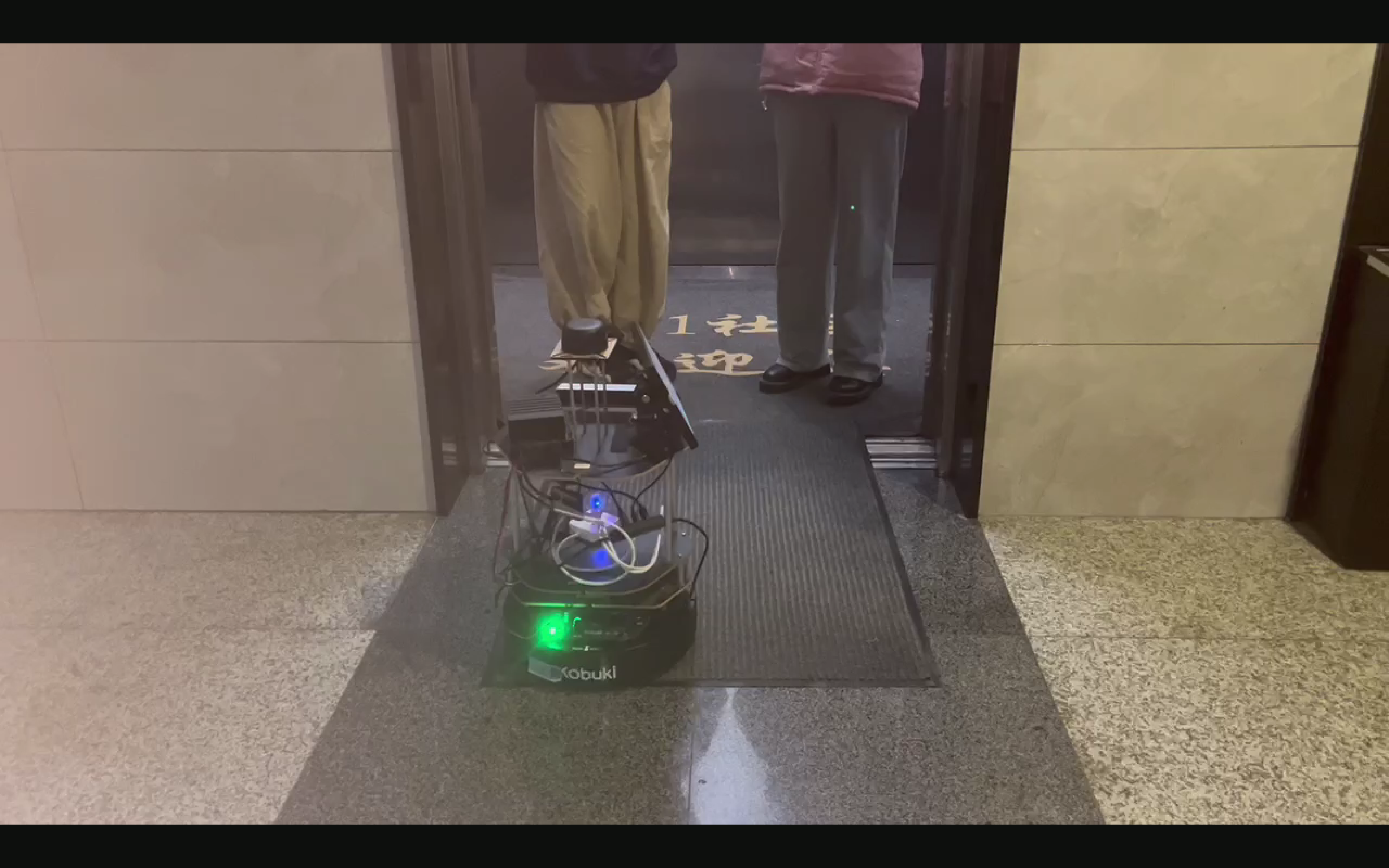}
	\caption{Humans inside the elevator block the door. The robot takes an active voice interaction.}
	\label{fig:beep1}
\end{figure}

\begin{figure}[htbp]
	\centering
	\includegraphics[width=0.75\linewidth]{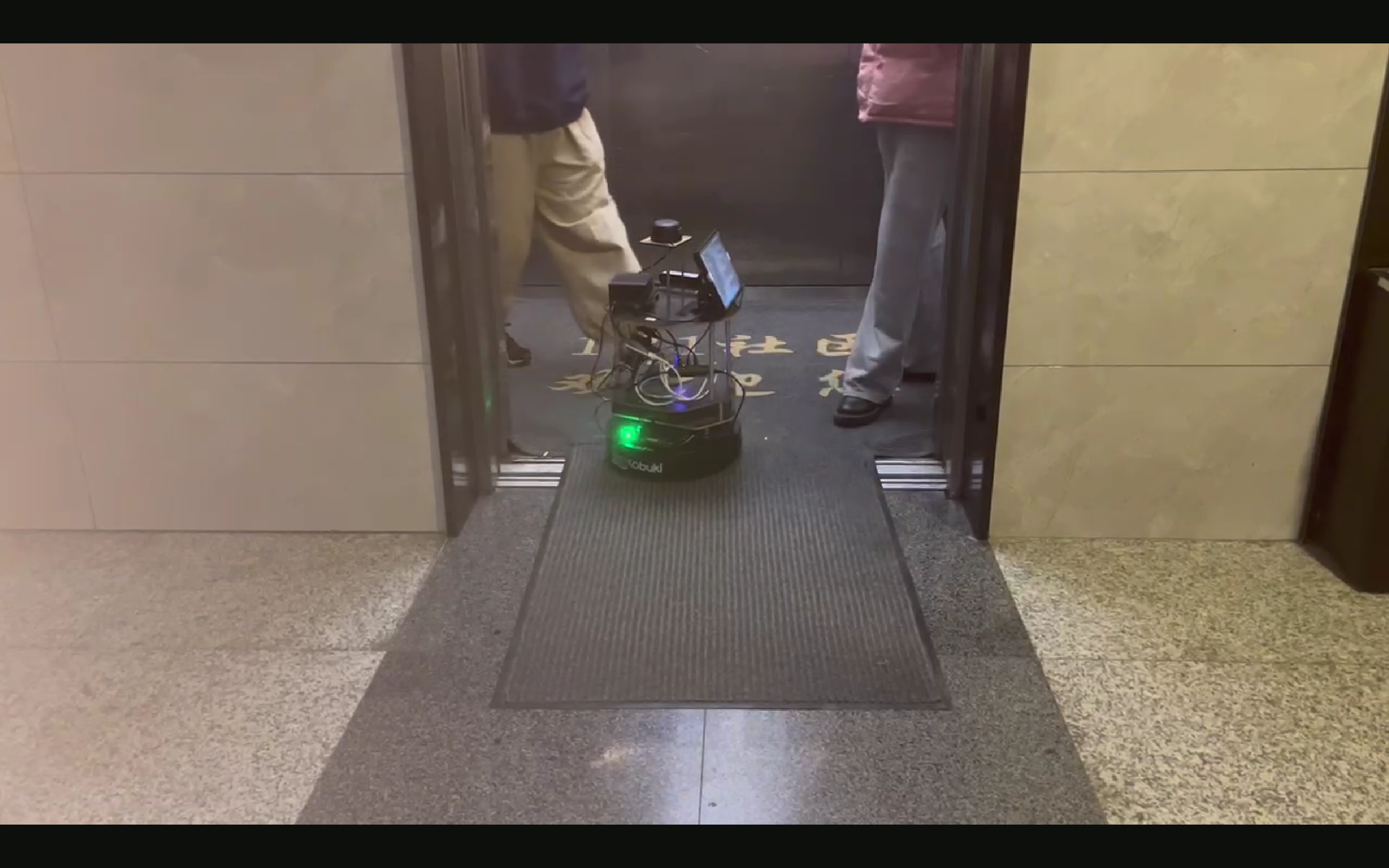}
	\caption{After the clear path operation, the robot enters the elevator successfully.}
	\label{fig:beep2}
\end{figure}

\textbf{(2)Humans are close to the elevator door}: There is plenty of room left for the robot to enter the elevator, but the current path of the robot is blocked. the real experiment results show that the proposed method takes an active voice interaction to clear the path, and then the robot enters the elevator when finding an available route. The figures of this process are demonstrated in Fig. \ref{fig:beep1} and Fig. \ref{fig:beep2}.

\section{Conclusion}

In this paper, we designed a reinforcement learning-based navigation algorithm working with active voice interaction. The proposed method reaches a high efficiency for the mobile robot on the basis of safety. Also, the proposed transform encoder-based NavFormer shows its advantages when calculating the state value, which is better than a state-of-the-art method SARL based on Fully Convolutional Network (FCN). Meanwhile, the real experiments also achieve the desired results. However, there still has some limitations, typically, the proposed method uses a discrete action space, which sometimes causes the robot runs not smoothly in real experiments. This problem could be optimized by using a suitable robot control algorithm or using the reinforcement learning method with a continuous action space, such as Deep Deterministic Policy Gradient. Our future work will focus on solving this problem.

\bibliographystyle{IEEEtran}

\vfil\vfil\vfil\vfil\vfil\vfil\vfil\vfil\vfil\vfil\vfil\vfil\vfil\vfil

\end{document}